\documentclass{ecai} 

\usepackage{latexsym}
\usepackage{amssymb}
\usepackage{amsmath}
\usepackage{amsthm}
\usepackage{booktabs}
\usepackage{enumitem}
\usepackage{graphicx}
\usepackage{color}
\usepackage{multirow}
\usepackage{bm}
\usepackage{arydshln}

\newcommand{\BibTeX}{B\kern-.05em{\sc i\kern-.025em b}\kern-.08em\TeX}
\usepackage{color,soul}
\usepackage{xcolor}
\definecolor{cerulean}{rgb}{0.0,0.48,0.65}
\definecolor{green}{rgb}{0.01, 0.75, 0.24}
\definecolor{Black}{RGB}{0.0, 0.0, 0.0}

\newcommand{\olive}[1]{\textcolor{Black}{#1}}
\newcommand{\br}[1]{\textcolor{Black}{#1}}

\newcommand{\grn}[1]{\textcolor{Black}{#1}}

\newcommand{\teal}[1]{\textcolor{Black}{#1}}
\newcommand{\brown}[1]{\textcolor{Black}{#1}}
\newcommand{\shadow}[1]{}
\def\o{\olive}
\def\s{\shadow}

\def\g{\grn}

\def\br{\brown}

\begin{document}


\begin{frontmatter}


\paperid{946} 


\title{Open-Set Multivariate Time-series Anomaly Detection}


\author[A,B]{\fnms{Thomas}~\snm{Lai}}
\author[A,B]{\fnms{Thi Kieu Khanh}~\snm{Ho}\thanks{Corresponding Author. Email: thi.k.ho@mcgill.ca.}}
\author[A,B]{\fnms{Narges}~\snm{Armanfard}} 

\address[A]{McGill University} 
\address[B]{Mila - Quebec AI Institute}


\begin{abstract}
Numerous methods for time-series anomaly detection (TSAD) have emerged in recent years, most of which are unsupervised and assume that only normal samples are available during the training phase, due to the challenge of obtaining abnormal data in real-world scenarios. Still, limited samples of abnormal data are often available, albeit they are far from representative of all possible anomalies. Supervised methods can be utilized to classify normal and seen anomalies, but they tend to overfit to the seen anomalies present during training, hence, they fail to generalize to unseen anomalies. We propose the first algorithm to address the open-set TSAD problem, called \textbf{\underline{M}}ultivariate \textbf{\underline{O}}pen-\textbf{\underline{S}}et time-series \textbf{\underline{A}}nomaly \textbf{\underline{D}}etector (MOSAD), that leverages only a few shots of labeled anomalies during the training phase in order to achieve superior anomaly detection performance compared to both supervised and unsupervised TSAD algorithms. MOSAD is a novel multi-head TSAD framework with a shared representation space and specialized heads, including the Generative head, the Discriminative head, and the Anomaly-Aware Contrastive head. The latter produces a superior representation space for anomaly detection compared to conventional supervised contrastive learning. Extensive experiments on three real-world datasets establish MOSAD as a new state-of-the-art in the TSAD field.
\end{abstract}

\end{frontmatter}


\section{Introduction}\label{sec:intro}
Multivariate time-series data is defined as time-ordered data where multiple variables are measured simultaneously at different time stamps or observations \cite{luo2018multivariate}. Each variable can represent an attribute or sensor within a sensory system. Analyzing multivariate time-series data is a complex task that necessitates careful considerations of several key factors, including temporal dependencies, dimensionality, non-stationarity and noise \cite{ho2023graph}. Time-series anomaly detection (TSAD) is the field of identifying patterns that deviate from the expected behavior within the time-series \cite{blazquez2021review,hojjati2022self}. These atypical patterns can manifest in various real-world applications such as sudden changes in transaction frequency, unusual network traffic, extreme weather events, or irregular vital signs related to the heart or brain.

Many \emph{unsupervised} deep learning-based methods have been proposed to automatically detect anomalies in time-series data \cite{zhao2020multivariate,shen2020timeseries,audibert2020usad,deng2021graph,ho2023self,hojjati2023multivariate}. These methods assume that only normal data is available in the training phase, since in practical scenarios, anomalies are rare events, hence, gathering an adequate quantity of labeled anomalies spanning all categories is very challenging. Thus, the general strategy of unsupervised methods is to exhaustively model the normal data such that in the test phase, anomalies would be detected if they significantly deviate from the learned normal patterns. Meanwhile, a small number of labeled anomaly samples are often accessible in many real-world datasets, such as the abnormal heart rhythm and brain activity confirmed by experienced readers of physiological signals, or the exceptionally high temperatures and unusual atmospheric conditions easily verified by environmental experts. These anomaly samples provide valuable information about application-specific abnormality \cite{liu2019margin}, yet unsupervised detectors have no mechanism that can take advantage of such information.

On the other hand, \emph{supervised} methods are able to learn from labeled anomaly data \cite{ijcai2022p394,tang2021self,ahmedt2020neural,pang2019deep}. Generally, they would assume that the available labeled abnormal data is comprehensive and represents well the distribution of all possible anomalies, i.e., these methods are binary classifiers with a normal and an abnormal class. However, it is highly impractical to assume that the available anomaly samples contain sufficient knowledge of all the anomaly distributions that will be encountered in the future. Additionally, the high imbalance between the normal and the abnormal classes would result in high false negative rates where anomalies are not accurately detected, because supervised methods are prone to overfitting to anomalies that resemble those encountered during training. This limitation arises from the substantial dissimilarity that can exist between seen and unseen anomaly classes \cite{ding2022catching}.

Recently, \textit{open-set} anomaly detection (OSAD) methods are proposed to leverage a limited number of anomalous samples in the training phase to achieve superior detection performance compared to unsupervised detectors, while not overfitting to those anomalies seen during training, as observed with supervised methods. The objective of open-set methods is to detect both seen anomalies encountered during training as well as unseen anomalies that are not presented during training. It is important to note that OSAD is different from the research fields of open-set recognition (OSR) \cite{bendale2016towards} and out-of-distribution detection (OOD) \cite{ren2019likelihood}. Specifically, OSAD learns a representation and defines a decision boundary about normal data, while leveraging few anomalies to push seen and unseen anomalies away from normal data. In contrast, OSR and OOD aim to learn the distribution of multiple classes \cite{mahdavi2021survey} with (assumed) large enough training data for each class, an assumption only made for the normal class in OSAD. 

Open-set TSAD remains an unaddressed challenge in the field. While several open-set algorithms have been proposed for image and video data \cite{ding2022catching,Zhang_2023_CVPR,Yao_2023_CVPR,tian_anomaly_2022}, previous studies have shown that by directly applying computer vision techniques to multivariate time-series data without modification, the results are sub-par compared to algorithms that are tailored for time-series data \cite{tang2021self,hojjati2022self,xu2021anomaly,hojjati2023multivariate}, due to the following reasons: First, the topology of images and videos is different from that of multi-sensor systems \cite{tang2021self}. Second, data augmentation techniques commonly used in computer vision, such as rotation and cropping, are not suitable for multivariate time-series data \cite{hojjati2022self}. Third, anomalies in time-series data exhibit unique characteristics and rarity patterns, different from image anomalies \cite{xu2021anomaly,ho2023multivariate}. Lastly, conventional methods for defining positive and negative pairs in contrastive learning that are widely used in computer vision cannot be directly employed for time-series data \cite{ho2023multivariate}. Therefore, to address this gap in the field, we propose the first open-set TSAD algorithm named \textbf{\underline{M}}ultivariate \textbf{\underline{O}}pen-\textbf{\underline{S}}et time-series \textbf{\underline{A}}nomaly \textbf{\underline{D}}etector (MOSAD). The main contributions of this paper are described as follows:

\begin{itemize}
    \item We propose MOSAD, trained in an open-set environment, where normal time-series and a small number of labeled anomalous time-series from a limited class of anomalies are used during training, with the goal of effectively detecting both seen and unseen anomaly classes -- a novel and unaddressed problem in the field. To the best of our knowledge, MOSAD marks the first study into open-set TSAD, which is challenging due to the high risk of overfitting to seen anomalies. 
    \item MOSAD encompasses three primary modules: the Feature Extractor, the Multi-head Network, and the Anomaly Scoring module. The Feature Extractor extracts features that are shared by the complementary Generative, the Discriminative, and the Anomaly-Aware Contrastive heads in the Multi-Head Network. The Anomaly Scoring module then combines the anomaly scores obtained from the three heads to detect anomalies.
    \item \g{We propose a novel Anomaly-Aware Contrastive Head in which, unlike vanilla supervised contrastive learning, where the similarity between samples of all classes are maximized, the similarity between seen anomaly samples is not maximized. This leads to superior anomaly detection performance due to the better representation space obtained.}
    \item We perform extensive experiments on three real-world multivariate time-series datasets obtained from industrial and biomedical domains under two experimental setups: the general and hard settings. In the general setting, the few labeled anomalies used in training are drawn from all possible anomaly classes, while in the hard setting, they are drawn from only one anomaly class. The results show that our approach consistently outperforms existing unsupervised and supervised methods, and thus establishes a new state-of-the-art in the TSAD field. 
\end{itemize}

\section{Proposed Method}
\begin{figure*}[!t]
\centering
\includegraphics[width=0.9\linewidth]{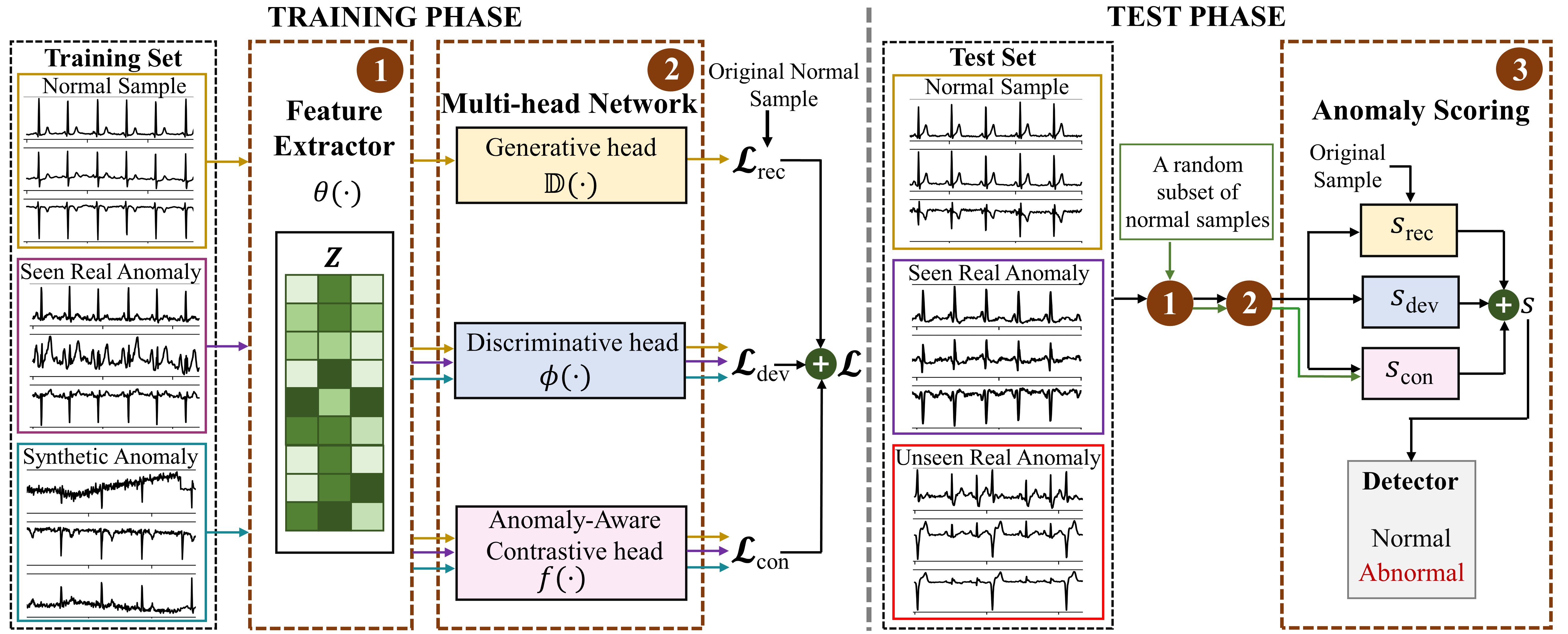}
\caption{The overall framework of MOSAD.}
\label{fig:model}
\end{figure*}

We define a multivariate time-series dataset for an open-set problem as $X = \{\mathbf{x}^{(i)}\}_{i=1}^{N+A},$ in which $X_n = \{\mathbf{x}^{(1)},\mathbf{x}^{(2)},\ldots,\mathbf{x}^{(N)})\}$ is the normal data and $X_a = \{\mathbf{x}^{(N+1)},\mathbf{x}^{(N+2)},\ldots,\mathbf{x}^{(N+A)}\}$ is the limited set of labeled anomalies $(A \ll N).$ $\mathbf{x}^{(i)} = \{x_k^{(i)}\}_{k=1}^K$ is the $i$th time interval\s{(i.e., observation)} in the dataset of $N+A$ time intervals, and $\mathbf{x}^{(i)} \in \mathbb{R}^{K \times L}.$ Each time interval $\mathbf{x}^{(i)}$ can be conceptualized as $L$ timestamps collected from $K$ variables (i.e., attributes/sensors) during this interval. Each variable, i.e., $x_k^{(i)},$ is represented as a vector in ${\mathbf{x}^{(i)}}$, where $x_k^{(i)} \in \mathbb{R}^{1 \times L}.$ Note that our objective is to detect anomalous time-series intervals. Intervals are also referred to as samples throughout the paper.

Our task is to develop a model that is trained on the training set $X$ where the limited set of labeled anomalies $X_a$ belongs to a set of seen anomaly classes $\mathcal{C} \subset \mathcal{S},$ where $\mathcal{S} = \{s_v\}_{v=1}^{|\mathcal{S}|}$ denotes the set of all possible anomaly classes, and $|\mathcal{S}|$ is the cardinality of $\mathcal{S}.$ Note that to mimic practical scenarios, we do not use the class labels of anomalies during training, even if they are given. Hence, in the test phase, we aim to detect anomalies spanning all anomaly classes. The Anomaly Scoring module is then designed to assign larger anomaly scores to both seen and unseen anomalies than normal samples. The block diagram of MOSAD, illustrating the three modules, namely Feature Extractor, Multi-head Network, and Anomaly Scoring, is shown in Figure \ref{fig:model}. Details of each module are presented below. 

\subsection{Feature Extractor} \label{sec:feature_extractor}
The foundational step in building a robust and interpretable open-set TSAD algorithm is to develop a feature learning backbone that can transform the raw multivariate time-series data into a more suitable format for the model, referred to as embeddings, that enables generalized and accurate anomaly detection in the subsequent stages. Various deep neural networks for time series data can be employed as feature extractors, such as long short-term memory (LSTM) \citep{karim2017lstm}, structured state space models \citep{ho2023multivariate}, and graph attention networks \citep{zhao2020multivariate}. \br{Notably, attention-based transformer networks have also been proposed to extract features from time series \citep{tang2021probabilistic,yang2023dcdetector,xu2021anomaly}.} However, for the sake of simplicity and to ensure a fair comparison with the current SOTA methods \citep{ijcai2022p394}, we utilize a temporal convolutional network (TCN) \citep{bai2018empirical} as our feature extractor, which is a widely recognized choice in the field capable of exploiting the \textit{temporal dependencies} existing in time-series data. Intuitively, we define $\mathbf{Z}^{(i)}$ as the feature embedding of $\mathbf{x}^{(i)}$ obtained from the feature extractor $\theta (\cdot).$ In other words, $\mathbf{Z}^{(i)} = \theta(\mathbf{x}^{(i)}),$ where $\mathbf{Z}^{(i)} \in \mathbb{R}^{1 \times D},$ and $D$ is the embedding dimension. 

\subsection{Multi-head Network}\label{sec:multi-head}
In this module, we propose a multi-head network of three downstream heads, namely Generative, Discriminative, and Anomaly-Aware Contrastive heads, where each head specializes in complementary aspects of open-set multivariate TSAD. Specifically, the Generative head uses generative learning to exhaustively model both the intricate temporal dependencies and \textit{inter-sensor dependencies} in normal data only, acquiring extensive knowledge of the normal class. The Discriminative head uses deviation learning to leverage a few seen anomalies to maximize the distance between both seen and unseen anomaly classes and the normal class, w.r.t. the decision boundary. Empirical results that prove the feasibility of combining generative and discriminative learning under a shared feature space for TSAD are shown in Section \ref{sec:exp_results}. The Anomaly-Aware Contrastive head then further strengthens the feature space under the paradigm of using contrastive representation learning to improve the performance of downstream tasks. For this head, we propose a new contrastive learning setup that is tailored to open-set TSAD, i.e., we maximize the similarity between normal samples, and we minimize the similarity between normal samples and seen anomaly types, but we do not maximize the similarity between anomalous samples, as we do not assume that anomalies belong to a single cluster in the representation space. Details of each of the heads are presented below.

\paragraph{Generative Head}\label{sec:gen_head} To learn the normal patterns within time-series data, we train the Generative head on only the subset of normal samples present in $X_n,$ and exclude the seen anomalies present in $X_a.$ We employ a generative approach based on the autoencoder framework that consists of an encoder and a decoder. The encoder is the feature extractor $\theta (\cdot)$ (i.e., the parameters are shared), while a decoder $\mathbb{D}(\cdot)$ with the reverse architecture of $\theta (\cdot)$ is added to reconstruct the time series. 

While we leverage temporal dependencies through the feature extractor described in Section \ref{sec:feature_extractor}, we also exploit inter-sensor dependencies inherent in multivariate time-series data by performing masked reconstruction. Specifically, we first define a masking function $M_{k}(\cdot),$ where $k=\{1,\ldots,K\}.$ $M_{k}(\mathbf{x}^{(i)})$ sets all values of the $k$th variable in $\mathbf{x}^{(i)}$ to 0. The task of the Generative head is to reconstruct the masked variable's feature vector, i.e., the zeroed $x_k^{(i)},$ with the information of the other unmasked variables.  This process is iterated across all $K$ variables. The reconstructions of the masked feature vectors of all variables are then combined to form the final reconstructed data, denoted as $\hat{\mathbf{x}}^{(i)},$ which can be represented as:
\begin{equation}\label{eq:xhati}
    \hat{\mathbf{x}}^{(i)} = \sum_{k=1}^K M'_k \Big(\mathbb{D}\Big(\theta \Big(M_k(\mathbf{x}^{(i)})\Big)\Big)\Big),
\end{equation}
where $M'_k(\cdot)$ zeros the values of other variables in the interval that are not the $k$th variable. The hypothesis is that the Generative head learns to reconstruct the masked feature vectors of normal samples during the training phase, while obtaining no expertise on reconstructing masked variables of abnormal samples, so the reconstruction loss of abnormal samples in the test phase would be high. The reconstruction loss is defined as:
\begin{equation} \label{eq:loss_rec}
    \mathcal{L}_{\text{rec}} = \| \hat{\mathbf{x}}^{(i)} - \mathbf{x}^{(i)} \|^2.
\end{equation} 

\paragraph{\g{Discriminative} Head} This head is designed to strengthen the decision boundary between normal data and anomalies by directly approximating the anomaly score. We use deviation learning to maximize the distance between normal and abnormal anomaly scores with only a few shots of seen anomaly samples and synthetic anomalies. Specifically, we employ a deviation function $dev(\cdot)$ that outputs a deviation score, referred to as an anomaly score, for the embedding $\mathbf{Z}^{(i)}$ of $\mathbf{x}^{(i)},$ which can be described as:
\begin{equation}
    dev (\mathbf{Z}^{(i)}) = \frac{\phi(\mathbf{Z}^{(i)}) - \mu}{\sigma},
\end{equation}
where $\phi(\cdot)$ is a neural network. $\mu$ and $\sigma,$ respectively, denote the mean and the standard deviation of the deviation scores of normal samples. We then perform deviation learning based on a Gaussian prior distribution \citep{pang2019deep}, in which $dev (\mathbf{Z}^{(i)})$ is imposed to approximate a score drawn from the prior if $\mathbf{x}^{(i)}$ is a normal sample. Hence, \g{following \citep{pang2019deep}}, we set $\mu$ and $\sigma$ to 0 and 1, respectively. Meanwhile, $dev (\mathbf{Z}^{(i)})$ is forced to have a statistically significant deviation from deviation scores of normal samples if $\mathbf{x}^{(i)}$ is anomalous. The deviation loss is then realized as:
\begin{equation} \label{eq:loss_dev}
    \mathcal{L}_{\text{dev}} = (1-y^{(i)})|dev(\mathbf{Z}^{(i)})| + y^{(i)} \text{max}(0,c-dev(\mathbf{Z}^{(i)})),
\end{equation}
where $|\cdot|$ denotes the absolute value, and $y^{(i)}$ represents the label of $\mathbf{x}^{(i)}.$ $y^{(i)} = 0$ if $\mathbf{x}^{(i)} \in X_n,$ otherwise, $y^{(i)} \in (0,1]$ if $\mathbf{x}^{(i)} \in \{X_a, X_a+\}.$ Note that $X_a+$ is a synthetic anomaly set generated by anomaly augmentation strategies, which do not attempt to characterize the full data distributions of anomalies but rather help MOSAD tackle the challenge of limited availability of real anomalies in the training phase. $c$ is denoted as a Z-score confidence interval hyperparameter. In general, $\mathcal{L}_{\text{dev}}$ enables MOSAD to push deviation scores of normal samples close to $\mu,$ while forcing a deviation for all seen anomalies of at least $c,$ where the value of $c$ is in between $\mu$ and the deviation scores of the anomalies. \g{Following \citep{pang2019deep}}, we set $c$ to 5 to achieve a high significance level for all seen anomalies.

It is worth mentioning that as we verify MOSAD's performance on both industrial and physiological signals, it is crucial to generate synthetic anomalies that can closely resemble real-world anomalies. Hence, we adopt two effective anomaly augmentation techniques, namely, contextual outlier exposure (COE) and window mixup (WMix) \citep{ijcai2022p394}. Regarding the COE method, to make an $\mathbf{x}^{(i)}$ become a synthetic anomaly, we randomly select a short window within $\mathbf{x}^{(i)}.$ Then, we select two variables in $\mathbf{x}^{(i)}$ and their corresponding data within the short window are swapped, and we set $y^{(i)} = 1.$ In regard to the WMix method, we generate a new anomaly sample $\mathbf{x}_\text{new},$ which is a combination of two existing time intervals $\mathbf{x}^{(i)}$ and $\mathbf{x}^{(j)},$ as $\mathbf{x}_{\text{new}}=\gamma \mathbf{x}^{(i)} + (1-\gamma)\mathbf{x}^{(j)},$ where $\gamma$ is sampled from the Beta distribution \citep{gupta2004handbook} as $\gamma \sim \text{Beta}(\alpha,\alpha).$ In our experiments, $\alpha$ is fixed to 0.05 \o{following \citep{ijcai2022p394}}. The corresponding new soft label is given by $y_{\text{new}}=\gamma y^{(i)} + (1-\gamma)y^{(j)},$ $y_{\text{new}} \in (0,1].$  

\paragraph{\g{Anomaly-Aware Contrastive Head}}\label{sec:contrastive} The Contrastive head is designed to create a latent representation space such that the similarity between normal samples is maximized while the similarity between normal and abnormal samples is minimized. Importantly, \g{unlike the vanilla supervised contrastive learning paradigm} that maximizes all the intra-class similarities \citep{khosla2020supervised,hojjati2022self}, we propose not to maximize the similarity between abnormal samples, as we do not assume that abnormal samples belong to a single cluster in the representation space; this \g{results in a representation space that is more appropriate for anomaly detection.}

Given $X,$ we define a mini-batch of size $\hat{N}$ as $\mathcal{B} = \{\mathbf{x}^{(1)},\mathbf{x}^{(2)},\ldots,\mathbf{x}^{(\hat{N})}\},$ $\hat{N} \ll N+A.$ Note that $\mathcal{B}$ can contain both normal and abnormal samples. Similar to the \g{Discriminative} head, we include both real and synthetic anomalies. We denote a normal sub-mini-batch $\mathcal{B}_n$ that consists of only normal samples, i.e.,  $\mathcal{B}_n = \mathcal{B}$\textbackslash$\{X_a\}.$ We first pass $\mathbf{Z}^{(i)}$ through the Contrastive head $f(\cdot),$ producing $\mathbf{G}^{(i)} = f(\mathbf{Z}^{(i)}),$ $\mathbf{G} \in \mathbb{R}^{1 \times d},$ where $d$ is the embedding dimension of the Contrastive head. The contrastive loss is then defined as:
\begin{equation} \label{eq:loss_con}
    \mathcal{L}_{\text{con}} = \sum_{i \in \mathcal{B}_n} \frac{-1}{|\mathcal{Q}_i|} \sum_{q \in \mathcal{Q}_i} \log \frac{\exp(\mathbf{G}_i.\mathbf{G}_q/\delta)}{\sum_{b \in \mathcal{B}_i}\exp(\mathbf{G}_i.\mathbf{G}_b/\delta)},
\end{equation}
where $\mathcal{B}_i = \mathcal{B}$\textbackslash$\{i\},$ $\mathcal{Q}_i=\mathcal{B}_n$\textbackslash$\{i\},$ and $\delta$ is the temperature hyperparameter to determine the strength of attraction or repulsion between representation vectors, \g{which is set to 0.07 following the implementation of \citep{khosla2020supervised}}.

The final loss in the training phase is the sum of the three losses presented in Equations \eqref{eq:loss_rec}, \eqref{eq:loss_dev}, \eqref{eq:loss_con}:
\begin{equation}
    \g{\mathcal{L} = \mathcal{L}_{\text{rec}} + \mathcal{L}_{\text{dev}} + \mathcal{L}_{\text{con}}.}
\end{equation}


\subsection{Anomaly Scoring} \label{sec:anomaly_scoring}

In the test phase, we compute the overall anomaly score $s$ of a test sample, i.e., $\mathbf{x}^{(t)},$ based on anomaly scores obtained from the three heads. More specifically, the Generative head produces an anomaly score $s_{\text{rec}}$ for each $\mathbf{x}^{(t)}$ by comparing $\mathbf{x}^{(t)}$ and $\hat{\mathbf{x}}^{(t)},$ which has been masked and reconstructed \teal{under the same procedure as} the training phase, i.e., $s_{\text{rec}} = \|\hat{\textbf{x}}^{(t)} - \textbf{x}^{(t)}\|^2.$ The underlying assumption is that normal samples would be well-reconstructed, while anomalies would be reconstructed poorly. \g{$s_{\text{rec}}$ is then normalized between 0 and 1 based on the minimum and maximum of a portion of the training set consisting of only normal samples, as the numeric range of $s_{\text{rec}}$ varies depending on the dataset.} On the other hand, the anomaly score $s_{\text{dev}}$ of $\mathbf{x}^{(t)}$ produced by the \g{Discriminative} head is the direct output of $\phi(\cdot),$ i.e., $s_{\text{dev}} = \phi(\mathbf{Z}^{(t)}).$ Despite the other losses being directly useful for anomaly scoring, using $\mathcal{L}_{con}$ would require all training data, both normal and abnormal, in the test phase, for its value to be meaningful, which is impractical. Instead, our experiments showed that the effect of the contrastive head on the latent space can be leveraged for anomaly scoring by comparing the similarity of a test sample with only a small subset of normal training data. To do this, the Contrastive head first predicts the normality of $\mathbf{x}^{(t)}$ by measuring the cosine similarity ($\text{Sim}\{\cdot\}$) \citep{rahutomo2012semantic} between $\mathbf{x}^{(t)}$ and a random set of normal samples $\mathcal{P}$ drawn from the training set, which is averaged across $\mathcal{P}.$ Then, the abnormal score $s_{\text{con}}$ is obtained by subtracting the score of normality from 1, i.e., $s_{\text{con}} = 1 - \frac{1}{\mathcal{|P|}}\sum_{p = 1, p \in \mathcal{P}}^{|\mathcal{P}|} \text{Sim}\{\mathbf{G}^{(t)},\mathbf{G}^{(p)}\},$ where $|\mathcal{P}|$ denotes the cardinality of $\mathcal{P}.$ By comparing $\mathbf{x}^{(t)}$ with a set of normal samples, both seen and unseen anomalies would result in a high \teal{$s_{\text{con}}$}. The final $s$ is thus computed as:
\begin{equation}\label{eq:anomaly_score}
     s = s_{\text{rec}} + s_{\text{dev}} + s_{\text{con}}.
\end{equation}
\g{It is important to note that in the TSAD research field, selecting a decision threshold on $s$ is not trivial, and it is common practice where researchers report their results with the best decision threshold on the \emph{labeled} test set \citep{ijcai2022p394}, leading to biased and unfair performance. To tackle this issue, and following the OSAD literature in the computer vision field \citep{ding2022catching}, we evaluate the performance of our proposed method using \emph{threshold-independent} metrics. In practice, a decision threshold can be selected based on the precision-recall trade-off.}

\subsection{Detecting Unseen Anomalies} 

\br{The ensemble nature of MOSAD enables the algorithm to detect unseen anomalies in addition to seen anomalies. The Generative head learns to reconstruct normal data only, thus, during the test phase, anomalous samples, both seen and unseen, would be reconstructed poorly. While the Contrastive head is trained with seen anomalies, its anomaly score is based on normal data only. Thus, any sample that is dissimilar to normal data, which includes both potential seen and unseen anomalies, would be scored as anomalies. Lastly, since all three anomaly scores would be 0 for a perfect normal sample, the combination of any positive terms in Equation \ref{eq:anomaly_score} will facilitate the detection of both seen and unseen anomalies.}

\section{Experiments}
\subsection{Experimental Settings}
In this section, we introduce the datasets, baselines, experimental settings, implementation details, and experimental results in our study.

\subsubsection{Datasets}
Numerous TSAD studies proposed the use of synthetic anomaly datasets, which are constructed with synthetic anomalies generated based on predefined assumptions. The rationale behind this approach is to create a clear difference between anomalies and normal samples, subsequently developing algorithms to align with these assumptions, thus improving the models' performance \citep{wang2021tsagen,lai2021revisiting,ijcai2022p394}. However, various actual anomalies in many real-world applications often exhibit subtle or small differences compared to normal samples. Thus, these generated anomalies may not fully capture the complexity and diversity of real-world anomalies. As a result, models trained extensively on synthetic anomalies often fail to generalize well to the real anomalies. On the other hand, a significant portion of existing TSAD studies have trained their models on widely used benchmark datasets such as Yahoo \citep{laptev2015s5}, NASA \citep{hundman2018detecting}, SWaT \citep{mathur2016swat}, WADI \citep{ahmed2017wadi}, SMAP \citep{su2019robust}. However, as highlighted in \cite{wu2021current,wagner2023timesead}, these benchmark datasets contain various flaws, including (i) mislabeled ground truth, (ii) triviality, (iii)
unrealistic anomaly density, and (iv) run-to-failure bias (prompting algorithms to simply detect the last points as anomalies) \cite{wu2021current}, and (v) distributional shift \cite{wagner2023timesead}, rendering them unfit for evaluating and comparing TSAD algorithms.

Knowing these issues, we carefully select the datasets that we use to validate the performance of our proposed method. First, we select the Server Machine Dataset (SMD) from the industrial domain to anchor our model onto a recogized dataset in the TSAD field. While not perfect, [2] showed that SMD is of much higher quality than the other criticized benchmarks. This study centers its focus on two additional publicly available datasets, namely, Physikalisch-Technische Bundesanstalt XL (PTB-XL) \citep{wagner2020ptb}, and Temple University Hospital Electroencephalogram Seizure Corpus (TUSZ) \citep{shah2018temple}. These datasets are well-established datasets in the biomedical domain and have not received criticisms (i)-(v). To ensure the quality of the labels, they are annotated by a consensus between a panel of 2-5 experts to minimize human errors and based on factors such as patient history, symptoms, diagnostic tests, and treatment outcomes.

More specifically, SMD is a 5-week-long dataset collected from different machines in a large Internet company with anomalies that are annotated by domain experts. PTB-XL is a large electrocardiography (ECG) dataset comprising of healthy and pathological heart rhythms. The latter includes four diagnostic classes, namely, myocardial infarction (MI), ST/T change (STTC), conduction disturbance (CD), and hypertrophy (HYP). TUSZ is the largest electroencephalogram (EEG) database, with normal data representing brain activity in resting-states and abnormal data representing seizures. This dataset contains three major anomaly class labels, namely, focal non-specific seizure (FNSZ), generalized non-specific seizure (GNSZ), and complex partial seizure (CPSZ). Note that the multiple anomaly classes in PTB-XL and TUSZ are important to our study for the construction of the open-set environment. Details of each dataset are shown in Table \ref{tab:datasets}.

\addtolength{\tabcolsep}{4pt}   
\begin{table}[h]
    \caption{\g{The three datasets used in our study. "Entities" denotes the number of machines in SMD or patients in PTB-XL and TUSZ. Sampling Rate is reported in Hertz. $K$ and $L$ are the number of variables and the length of a time interval, respectively.}}
    \centering
    \scalebox{0.95}{
    \begin{tabular}{cccccc}
        \hline
         & \textbf{Dataset} & \textbf{Entities} & \textbf{Sampling Rate} & \textbf{$K$} & \textbf{$L$} \\ \hline \hline
         & SMD & 28 & $\frac{1}{60}$ & 38 & 200 \\
         & PTX-XL & 18,885 & 100 & 12 & 1000 \\
         & TUSZ & 43 & 100 & 19 & 1000 \\ \hline
    \end{tabular}}
    \label{tab:datasets}
\end{table}

\g{In order to set up the open-set environment that mimics real-world applications \citep{ding2022catching}, where the number of anomaly samples available is much smaller than that of normal samples, only a limited number of real anomaly samples, denoted as $\eta,$ are used in the training phase for the open-set experimental setup. The value of $\eta$ is \s{arbitrarily} chosen such that each anomaly class represents about $0.1\%$ of the training set. The values of $\eta$ are shown in Table \ref{tab:data_division}.}

\g{It is also important to highlight that in addition to reducing the number of anomalous training samples according to the OSAD problem definition, our data preprocessing differs from previous papers in other ways. For instance, we employ the recent standard interval-level TSAD \citep{ijcai2022p394,ho2023self}, as opposed to point-level anomaly detection that enables point-adjustment bias, which is a common issue in TSAD \citep{wagner2023timesead}. Furthermore, in the SMD dataset, all 28 machines are combined and modeled together instead of individual machines modeled separately, as suggested by \citep{su2019robust}.}

\begin{table}[h]
    \centering
    \caption{\g{The numbers of \texttt{normal} / \texttt{real anomaly} samples in the training and test sets of each dataset.}}
    \scalebox{0.95}{
    \begin{tabular}{ccccc}
        \hline 
         & \textbf{Dataset} & \textbf{Setting} & \textbf{Training} & \textbf{Test} \\ \hline \hline
         & \multirow{2}{*}{SMD} & \textbf{U} & 5958 / 0  & \multirow{2}{*}{640 / 88} \\
         &  & \textbf{O}$_G$ & 5958 / 10  & \\\hline
         & \multirow{3}{*}{PTB-XL} & \textbf{U} & 8157 / 0  & \multirow{3}{*}{912 / 1286} \\
         & & \textbf{O}$_G$ & 8157 / 40 & \\
         & & \textbf{O}$_H^{\text{Class}}$ & 8157 / 10 & \\ \hline 
         & \multirow{3}{*}{TUSZ} & \textbf{U} & 39084 / 0 & \multirow{3}{*}{4343 / 316} \\ 
         &  & \textbf{O}$_G$ & 39084 / 90 &  \\ 
         &  & \textbf{O}$_H^{\text{Class}}$ & 39084 / 30 &  \\ \hline 
    \end{tabular}}
    \label{tab:data_division}
\end{table}

\subsubsection{Baselines} 

We compare MOSAD against deep learning-based methods in the TSAD literature based on two branches of experimental settings:  Unsupervised Setting (aka \textbf{U}) and Open-set Setting. For \textbf{U}, where only normal data is available in the training phase, we select the unsupervised methods LSTM-AE \citep{malhotra2016lstm}, TCN-AE \citep{THILL2021107751}, MTAD-GAT \citep{zhao2020multivariate}, THOC \citep{shen2020timeseries}, USAD \citep{audibert2020usad}, DVGCRN \citep{chen2022deep}, and DCdetector \cite{yang2023dcdetector} for comparison.

For the open-set setting, we establish two sub-settings, namely the General Setting (\textbf{O}$_G$) and the Hard Setting (\textbf{O}$_H^{\text{Class}}$), where $\text{Class}$ denotes the anomaly class visible in training. The general setting is a scenario where few labeled anomalies from all possible anomaly classes are present during training, i.e., $\mathcal{C} = \mathcal{S}.$ Note that the labeled anomalies used in training do not come from the test set. Since MOSAD is the first open-set TSAD algorithm, we also compare MOSAD with supervised binary-classifier baselines that utilize the cross-entropy loss \citep{zhang2018generalized}  such as LSTM \citep{karim2017lstm}, CNN-LSTM \citep{ahmedt2020neural}, GAT \citep{veličković2018graph}, and NCAD \citep{ijcai2022p394} in \textbf{O}$_G.$ Meanwhile, the hard setting is the most challenging scenario where few labeled anomalies from only one anomaly class are present during training (i.e., $|\mathcal{C}|=1,$ where $|\mathcal{C}|$ is the cardinality of $\mathcal{C}$), with the aim of detecting anomalies from all anomaly classes in the test phase. Since the hard setting requires a dataset with more than one distinct anomaly class, we evaluate the performance of MOSAD on PTB-XL \g{and TUSZ}, where MOSAD is compared with the binary classifiers LSTM, CNN-LSTM, GAT, TCN \citep{bai2018empirical}, and NCAD. Training and test sets of each dataset are illustrated in Table \ref{tab:data_division}. Note that identical datasets (with the same number and index of normal and abnormal samples) for both training and test phases are used for MOSAD and the comparison methods.

\subsubsection{Implementation Details}\label{section:implement}

The whole MOSAD network, including the Feature Extractor and the Multi-Head Network, is jointly optimized using the Adam optimizer \citep{KingBa15} with AMSGrad \citep{j.2018on}, initial learning rate of 1e-3, \g{batch size of 64,} and weight decay of 1e-5, to a maximum epoch of 30. The loss on a portion of the training set is used for early stopping. $\phi(\cdot)$ is a network encompassing two linear layers, a PReLU layer, a batch norm layer, and a dropout layer. $f(\cdot)$ is a fully connected layer followed by normalization. The same TCN architecture is used for MOSAD and all TCN-based comparison methods. All experiments are performed on a fixed random seed of 123. Other hyperparameters are $D=120,$ $d=32,$ and $|\mathcal{P}|=64.$ To assess the performance of all methods, we employ threshold-independent evaluation metrics, as mentioned in Section \ref{sec:anomaly_scoring}. We evaluate the models on Area Under the Receiver-Operating Characteristic Curve (AUC) following OSAD literature \citep{ding2022catching,Zhang_2023_CVPR,Yao_2023_CVPR}, and Area Under the Precision-Recall Curve (APR) which is commonly used when the classes are imbalanced \citep{davis2006relationship}.

\subsection{Experimental Results}\label{sec:exp_results}
\subsubsection{General Setting}

\begin{table}[t]
    \centering
    \caption{\g{Comparative analysis of our proposed MOSAD and existing methods under the unsupervised (\textbf{U}) and open-set general (\textbf{O}$_G$) settings. The best and second-best scores are respectively marked in bold and with $^{\star}.$}}
    \scalebox{0.95}{
    \begin{tabular}{cc|cccccc}
        \hline \hline
        \multicolumn{2}{c|}{\multirow{2}{*}{\textbf{Method}}} & \multicolumn{2}{c}{\textbf{SMD}} & \multicolumn{2}{c}{\textbf{PTB-XL}} & \multicolumn{2}{c}{\textbf{TUSZ}} \\ \cmidrule(lr){3-4} \cmidrule(lr){5-6} \cmidrule(lr){7-8}
         & & AUC & APR & AUC & APR & AUC & APR \\ \hline
         \multirow{7}{*}{\rotatebox[origin=c]{90}{Unsupervised (\textbf{U})}} & LSTM-AE & 51.4 & 13.3 & 63.4 & 74.5 & 63.2 & 10.6 \\ \cdashline{2-8}
         & TCN-AE & 66.3 & 20.2 & 63.7 & 74.4 & 72.1 & 11.6 \\ \cdashline{2-8}
         & MTAD-GAT & 50.6 & 19.7 & 64.5 & 68.8 & 69.0 &  10.8 \\ \cdashline{2-8}
         & THOC & 55.7 & 14.8 & 47.4 & 58.0 & 53.4 & 7.2  \\\cdashline{2-8}
         & USAD & 61.3 & 20.0 & 62.4 & 72.6 & 63.1 & 9.8  \\ \cdashline{2-8}
         %
         %
         & DVGCRN & 59.1 & 21.0 & 62.0 & 70.7 & 0.0 & 0.0  \\ \cdashline{2-8} 
         & DCdetector & 64.7 & 26.3 & 41.7 & 54.6 & 50.0 & 7.2 \\ \hline \hline
         \multirow{7}{*}{\rotatebox[origin=c]{90}{Open-Set (\textbf{O}$_G$)}} & LSTM & 50.1 & 21.0 & 63.8 & 74.2 & 66.2 & 10.8\\ \cdashline{2-8}
         & CNN-LSTM & 52.3 & 13.5 & 50.8 & 60.6 & 49.9 & 6.8 \\ \cdashline{2-8}
         & GAT & 53.4 & 28.5 & 57.0 & 65.6 & 62.8 & 11.7 \\ \cdashline{2-8}
         & NCAD & 63.1 & 31.6 & 60.8 & 71.8 & 67.8 & 27.4  \\ \cline{2-8}
         & MOSAD$_{\text{VSC}}$ & 69.2 & 31.7 & 67.7 & 77.4 & 70.9 & 30.2 \\ \cdashline{2-8}
        & MOSAD$_{\text{NMR}}$ & 69.5$^{\star}$ & 32.7$^{\star}$ & $\bm{69.7}$ & 77.4$^{\star}$ & 76.0$^{\star}$ & 48.4$^{\star}$ \\ \cdashline{2-8}
         & \textbf{MOSAD} & $\bm{70.0}$ & $\bm{41.1}$ & 69.0$^{\star}$ & $\bm{78.1}$ & $\bm{78.0}$ & $\bm{51.0}$ \\ \hline \hline
    \end{tabular}}
    \label{tab:result_general}
\end{table}

\begin{table*}[h]
\centering
\caption{\g{Comparative analysis of our proposed MOSAD and other methods on PTB-XL and TUSZ under the open-set hard settings (\textbf{O}$^{\text{Class}}_H$). The best and second-best scores are respectively marked in bold and with $^{\star}.$}}
\scalebox{0.95}{
\begin{tabular}{c|cccccccccc|cccccccc} 
    \hline \hline
     \multirow{3}{*}{\textbf{Method}} & \multicolumn{10}{c|}{\textbf{PTB-XL}} & \multicolumn{8}{c}{\textbf{TUSZ}}\\
     & \multicolumn{2}{c}{\textbf{O}$_H^{\text{MI}}$} & \multicolumn{2}{c}{\textbf{O}$_H^{\text{STTC}}$} & \multicolumn{2}{c}{\textbf{O}$_H^{\text{CD}}$} & \multicolumn{2}{c}{\textbf{O}$_H^{\text{HYP}}$} & \multicolumn{2}{c|}{\textbf{Average}} & \multicolumn{2}{c}{\textbf{O}$_H^{\text{FNSZ}}$} & \multicolumn{2}{c}{\textbf{O}$_H^{\text{GNSZ}}$} & \multicolumn{2}{c}{\textbf{O}$_H^{\text{CPSZ}}$} & \multicolumn{2}{c}{\textbf{Average}} \\ \cmidrule(lr){2-3} \cmidrule(lr){4-5} \cmidrule(lr){6-7} \cmidrule(lr){8-9} \cmidrule(lr){10-11} \cmidrule(lr){12-13} \cmidrule(lr){14-15} \cmidrule(lr){16-17} \cmidrule(lr){18-19} 
     & AUC & APR & AUC & APR & AUC & APR & AUC & APR & AUC & APR & AUC & APR & AUC & APR & AUC & APR & AUC & APR \\ \hline
     LSTM & 54.3 & 66.5 & 56.5 & 68.5$^{\star}$ & 56.9 & 70.3 & 43.6 & 57.2 & 52.8 & 65.6 & 61.3$^{\star}$ & 18.9$^{\star}$ & 64.3 & 29.7 & 67.0$^{\star}$ & 10.9 & 64.2$^{\star}$ & 19.8 \\ \hdashline
     CNN-LSTM & 55.0 & 64.5 & 48.6 & 58.4 & 50.1 & 58.9 & 49.9 & 58.0 & 50.9 & 60.0 & 52.6 & 8.2 & 49.8 & 7.4 & 51.7 & 7.9 & 51.4 & 7.8\\ \hdashline
     GAT & 56.8$^{\star}$ & 64.3 & 51.7 & 60.1 & 53.6 & 63.3 & 54.4 & 63.4 & 54.1 & 62.8 & 56.7 & 10.7 & 55.3 & 10.9 & 55.0 & 10.2 & 55.7 & 10.6\\ \hdashline
     TCN & 52.3 & 65.0 & 50.9 & 59.2 & 60.2 & 69.8 & 58.3 & 65.1 & 55.4 & 64.8 & 56.4 & 12.4 & 70.4$^{\star}$ & 35.3$^{\star}$ & 59.8 & 13.5$^{\star}$ & 62.2 & 20.4$^{\star}$\\ \hdashline
     NCAD & 56.2 & 69.2$^{\star}$ & 58.6$^{\star}$ & 67.4 & 63.4$^{\star}$ & 71.7$^{\star}$ & 59.1$^{\star}$ & 65.6$^{\star}$ & 59.3$^{\star}$ & 68.5$^{\star}$ & 49.7 & 8.9 & 59.7 & 19.3 & 51.4 & 12.8 & 53.6 & 13.7 \\ \hline
     \textbf{MOSAD} & $\bm{67.4}$ & $\bm{77.0}$ & $\bm{69.2}$ & $\bm{79.0}$ & $\bm{70.0}$ & $\bm{79.9}$ & $\bm{68.3}$ & $\bm{77.3}$ & $\bm{68.7}$ & $\bm{78.3}$ & $\bm{68.3}$ & $\bm{25.2}$ & $\bm{75.6}$ & $\bm{35.5}$ & $\bm{68.6}$ & $\bm{14.3}$ & $\bm{70.8}$ & $\bm{25.0}$ \\ \hline \hline
\end{tabular}}
\label{tab:result_hard}
\end{table*}

Table \ref{tab:result_general} assesses the performance of the methods under \textbf{U} and \textbf{O}$_G,$ which reveals that MOSAD consistently outperforms all SOTA methods in AUC and \o{APR} across the three datasets from diverse application domains. It is important to note that having access to even a limited number of seen anomalies during training provides MOSAD and supervised methods with an advantage over unsupervised baselines. However, under \textbf{O}$_G,$ the limited knowledge of seen anomalies during training causes the supervised binary classifier methods to overfit to seen anomalies, diminishing their ability to generalize to the entire anomaly distribution. In \g{most} cases, they are even  outperformed by the unsupervised methods. Notably, by surpassing the performance of unsupervised methods, MOSAD underscores its ability to retain unsupervised baseline performance with its Generative head, and by also outperforming supervised methods, MOSAD demonstrates its capability of leveraging the limited knowledge of anomalies given in training with its \g{Discriminative} and \g{Anomaly-Aware} Contrastive heads. 

To demonstrate the superiority of our proposed Anomaly-Aware Contrastive head described in Section \ref{sec:contrastive}, we include a variation of MOSAD, namely MOSAD$_{\text{VSC}},$ in Table \ref{tab:result_general}. Specifically, MOSAD$_{\text{VSC}}$ employs the vanilla supervised contrastive learning paradigm \citep{khosla2020supervised}, i.e., the similarity between abnormal samples are maximized during training. It is shown that MOSAD with our proposed Anomaly-Aware Contrastive head achieves higher performance in both AUC and APR compared to MOSAD$_{\text{VSC}}.$

To demonstrate the effectiveness of the masked reconstruction described in Section \ref{sec:gen_head}, we also implement MOSAD without masked reconstruction, denoted as MOSAD$_{\text{NMR}}$, as shown in Table \ref{tab:result_general}. In other words, MOSAD$_{\text{NMR}}$ does not explicitly model inter-sensor dependencies. Note that in MOSAD${_\text{NMR}}$, we employ a vanilla autoencoder framework for the Generative head, using the same encoder and decoder as the full MOSAD. The overall superior performance of MOSAD compared to MOSAD${_\text{NMR}}$ highlights the importance of explicitly modeling inter-sensor dependencies in multivariate time-series datasets.

\subsubsection{Hard Setting}
Table \ref{tab:result_hard} shows a comparison between the existing supervised methods and MOSAD on the PTB-XL \g{and TUSZ datasets} under \textbf{O}$^{\text{Class}}_H.$ As expected, the performance of MOSAD and comparison methods are lower under \textbf{O}$^{\text{Class}}_H$ than under \textbf{O}$_G,$ as less knowledge about anomalies is introduced during training under \textbf{O}$^{\text{Class}}_H,$ yet MOSAD outperforms all supervised methods in AUC and APR. More particularly, MOSAD achieves an increase of 9.4\% in mean AUC and 9.8\% in mean APR compared to the latest and most comparable algorithm NCAD on the PTB-XL dataset\br{\footnote{p<0.05 for both AUC and APR, according to the Wilcoxon signed-rank test \cite{siegel1956nonparametric}.}}. On the TUSZ dataset, MOSAD also achieves significant improvement compared to other methods, although MOSAD's APR performance varies greatly with the seen class, which shows the significance of the anomaly samples presented to the models during training. \s{Overall, the results in Table \ref{tab:result_hard} show that existing SOTA supervised TSAD algorithms are insufficient solutions to the open-set TSAD problem.}

\subsubsection{Ablation Study}
The following sections are dedicated to hyperparameter sensitivity and ablation studies on PTB-XL, which is the most challenging dataset, as it contains the most anomaly classes among the three datasets. 

 In this section, we conduct ablation studies that specifically focus on the Multi-head Network and the Anomaly Scoring Module, since, as mentioned in Section \ref{sec:feature_extractor}, we fix the Feature Extractor TCN because it is a commonly recognized choice in the field to ensure a fair and direct comparison with other methods. Particularly, seven ablations of MOSAD under \textbf{O}$^{\text{Class}}_H$ on PTB-XL are presented in Table \ref{tab:abl_individual_heads}. Each ablation consists of the Feature Extractor, and a Multi-Head Network and an Anomaly Scoring Module that contain only losses listed in the first column of Table \ref{tab:abl_individual_heads} during training and the anomaly scores listed in the second column during inference. \s{For example, \texttt{Gen+Con} contains the Generative and the Anomaly-Aware Contrastive heads in the Multi-Head Network and $s_\text{rec}$ and $s_\text{con}$ in the Anomaly Scoring Module. The results reveal that having only $\mathcal{L}_{\text{rec}}$ and $s_{\text{rec}}$} is the best \br{overall} single-head ablation, and that combining all three heads yields the best results. The results reveal that the removal of either $\mathcal{L}_{\text{rec}}$, $\mathcal{L}_{\text{con}}$, or $\mathcal{L}_{\text{dev}}$ resulted in significant drops in performance. The results also show that each individual head is effective on its own, and that the inclusion of all three losses and anomaly scores resulted in the best overall performance.
 
\begin{table}[ht]
    \centering
    \caption{Performance of MOSAD ablations w.r.t. individual heads and their combinations under \textbf{O}$_H^{\text{Class}}$ on PTB-XL. \br{The best and second-best scores are respectively marked in bold and with $^{\star}.$}}
    \scalebox{0.87}{
    \begin{tabular}{cc|cccccccc}
    \hline \hline
          \multicolumn{2}{c}{\textbf{MOSAD}} & \multicolumn{2}{c}{\textbf{O}$_H^{\text{MI}}$} & \multicolumn{2}{c}{\textbf{O}$_H^{\text{STTC}}$} & \multicolumn{2}{c}{\textbf{O}$_H^{\text{CD}}$} & \multicolumn{2}{c}{\textbf{O}$_H^{\text{HYP}}$} \\  \cmidrule(lr){1-2} \cmidrule(lr){3-4} \cmidrule(lr){5-6} \cmidrule(lr){7-8} \cmidrule(lr){9-10}
         Training & Inference & AUC & APR & AUC & APR & AUC & APR & AUC & APR\\ \hline
         $\mathcal{L}_{\text{rec}}$ & $s_{\text{rec}}$ & 61.0 & 73.0 & 65.9 & 74.9 & 65.6 & 73.3 & 66.4 & 74.0\\ \hdashline
         $\mathcal{L}_{\text{dev}}$ & $s_{\text{dev}}$ & 56.5 & 68.6 & 54.2 & 65.0 & 61.0 & 71.8 & 55.0 & 64.9 \\ \hdashline
         $\mathcal{L}_{\text{con}}$ & $s_{\text{con}}$ &  63.9 & 74.2 & 59.6 & 66.7 & 63.5 & 73.6 & 57.3 & 63.9 \\ \hdashline
         $\mathcal{L}_{\text{dev}},\mathcal{L}_{\text{con}}$ & $s_{\text{dev}},s_{\text{con}}$ &  57.8 & 69.3 & 55.8 & 65.9 & 58.3 & 70.4 & 54.3 & 61.8  \\ \hdashline
         $\mathcal{L}_{\text{rec}},\mathcal{L}_{\text{dev}}$ & $s_{\text{rec}},s_{\text{dev}}$ & 64.8 & 74.4 & 63.4 & 71.8 & 69.1 & 78.8 & 65.7 & 74.6  \\ \hdashline
         $\mathcal{L}_{\text{rec}},\mathcal{L}_{\text{con}}$ & $s_{\text{rec}},s_{\text{con}}$ & 65.0$^{\star}$ & 74.7$^{\star}$ & 68.0 & 78.2$^{\star}$ & 66.8 & 77.4 & 66.7$^{\star}$ & $\bm{77.3}$  \\ \hdashline
         All $\mathcal{L}$ & $s_{\text{rec}},s_{\text{dev}}$ & $\bm{67.4}$ & $\bm{77.0}$ & 69.1$^{\star}$ & $\bm{79.0}$ & 69.9$^{\star}$ & 79.7$^{\star}$ & 68.2$^{\star}$ & $\bm{77.3}$ \\  \hdashline
         All $\mathcal{L}$ & All $s$ & $\bm{67.4}$ & $\bm{77.0}$ & $\bm{69.2}$ & $\bm{79.0}$ & $\bm{70.0}$ & $\bm{79.9}$ & $\bm{68.3}$ & $\bm{77.3}$ \\ \hline \hline
    \end{tabular}}
    \label{tab:abl_individual_heads}
\end{table}

\subsubsection{Detection of Seen Versus Unseen Anomalies}\label{sec:seen_unseen} \g{Table \ref{tab:result_seen_unseen} shows the performance of MOSAD and NCAD on the Seen, Unseen, and Normal sample classes separately in the test phase under \textbf{O}$^{\text{Class}}_H$ on PTB-XL. \s{, as opposed to the results in Table \ref{tab:result_hard} that presents the performance on Seen and Unseen samples combined.}\br{In this experiment, only anomalies from the Seen class and normal samples are included during testing for the "Seen" column. Likewise, only anomalies that are not part of the Seen class and normal samples are included for the "Unseen" column. Lastly, the "Normal" column reports AUC and APR when both the labels and anomaly scores are subtracted from 1.} MOSAD consistently outperforms NCAD in detecting both sample types, except for the slight underperformance on the Seen class under \textbf{O}$^{\text{STTC}}_H.$ Despite this, the large margin of detection performance on the Unseen class of MOSAD over NCAD leads to a large difference in their overall performance, which shows the superior generalizability of MOSAD over NCAD.}

\begin{table}[h]
    \centering
    \caption{\g{Performance of MOSAD and NCAD w.r.t sample types (i.e., Seen anomalies, Unseen anomalies, and Normal) during testing under \textbf{O}$_H^{\text{Class}}$ on PTB-XL.} \br{The best scores are marked in bold.}}
    \scalebox{0.95}{
    \begin{tabular}{cc|cccccc}
    \hline \hline
         \multirow{2}{*}{\textbf{Setting}} & \multirow{2}{*}{\textbf{Method}} & \multicolumn{2}{c}{\textbf{Seen}} & \multicolumn{2}{c}{\textbf{Unseen}} & \multicolumn{2}{c}{\textbf{Normal}} \\ \cmidrule(lr){3-4} \cmidrule(lr){5-6} \cmidrule(lr){7-8}
         & & AUC & APR & AUC & APR & AUC & APR \\ \hline 
         \multirow{2}{*}{\textbf{O}$_H^{\text{MI}}$} & NCAD & 59.6 & 37.5 & 55.4 & 64.2 & 56.2 & 43.1 \\ \cdashline{2-8}
         & \textbf{MOSAD} & $\bm{64.9}$ & $\bm{43.8}$ & $\bm{67.9}$ & $\bm{73.6}$ & $\bm{67.4}$ & $\bm{53.0}$ \\ \hline 
         \multirow{2}{*}{\textbf{O}$_H^{\text{STTC}}$} & NCAD & $\bm{55.2}$ & $\bm{25.5}$ & 59.4 & 63.7 & 58.6 & 47.4 \\ \cdashline{2-8}
         & \textbf{MOSAD} & 51.0 & 24.2 & $\bm{73.4}$ & $\bm{79.2}$ & $\bm{69.2}$ & $\bm{54.8}$ \\ \hline 
         \multirow{2}{*}{\textbf{O}$_H^{\text{CD}}$} & NCAD & 67.5 & 35.4 & 62.7 & 67.9 & 63.4 & 51.3 \\ \cdashline{2-8}
         & \textbf{MOSAD} & $\bm{82.7}$ & $\bm{67.3}$ & $\bm{67.8}$ & $\bm{75.4}$ & $\bm{70.0}$ & $\bm{55.6}$  \\ \hline 
         \multirow{2}{*}{\textbf{O}$_H^{\text{HYP}}$} & NCAD & 48.4 & 5.6 & 59.6 & 65.5 & 59.1 & 48.3 \\ \cdashline{2-8}
         & \textbf{MOSAD} & $\bm{65.8}$ & $\bm{11.2}$ & $\bm{68.4}$ & $\bm{76.8}$ & $\bm{68.3}$ & $\bm{54.0}$ \\ \hline \hline
    \end{tabular}}
\label{tab:result_seen_unseen}
\end{table}

\subsubsection{Sensitivity Analysis}

In this section, we investigate the sensitivity of MOSAD by conducting two experiments on PTB-XL under the difficult setting, i.e., \textbf{O}$^\text{Class}_H,$ including : (1) varying the number of labeled anomalies $\eta$ used for loss minimization, and (2) varying the number of drawn normal samples $|\mathcal{P}|$ used for computing $s_{\text{con}}.$

\textbf{Effect of $\eta.$} Table \ref{tab:result_eta} shows a comparison between MOSAD and its SOTA competitor, i.e., NCAD, w.r.t. $\eta$ under \textbf{O}$^{\text{Class}}_H.$ The models were trained with different $\eta\in\{3,6,...,30\}$. The $\eta$ samples are selected such that, for example, $\eta=6$ includes all the samples in $\eta=3.$ The results are illustrated by calculating the Mean and Standard Deviation (STD) of the models' performance over the lower $\eta$ values and the higher $\eta$ values, respectively. The results reveal that MOSAD consistently outperforms NCAD, and that the mean AUC and APR of NCAD drastically decrease when $\eta\in\{3,6,9,12,15 \}$ versus when $\eta\in\{18,21,24,27,30 \}$, whereas MOSAD exhibits greater stability across a range of $\eta$. This shows the importance of the mechanisms in MOSAD that explicitly address the open-set problem, while suggesting that supervised methods like NCAD tend to overfit to seen anomalies, making them less effective at detecting various anomalies that are unseen during training.

\begin{table}[h]
\caption{Performance of MOSAD and NCAD w.r.t. in different $\eta$ ranges under \textbf{O}$_H^{\text{Class}}$ on PTB-XL, reported in terms of Mean and Standard Deviation (STD). \br{The best scores are marked in bold.}}
\centering
\scalebox{0.95}{
\begin{tabular}{cc|cccccccc}
    \hline \hline
     & \multirow{2}{*}{\textbf{Method}} & \multicolumn{2}{c}{\textbf{O}$_H^{\text{MI}}$} & \multicolumn{2}{c}{\textbf{O}$_H^{\text{STTC}}$} & \multicolumn{2}{c}{\textbf{O}$_H^{\text{CD}}$} & \multicolumn{2}{c}{\textbf{O}$_H^{\text{HYP}}$} \\ \cmidrule(lr){3-4} \cmidrule(lr){5-6} \cmidrule(lr){7-8} \cmidrule(lr){9-10}
     & & AUC & APR & AUC & APR & AUC & APR & AUC & APR\\ \hline
       \multicolumn{10}{c}{$\eta \in \{3,6,9,12,15 \}$} \\ \hline
\multirow{2}{*}{Mean} & NCAD  & 59.8 & 69.8   & 60.7 & 69.9     & 58.1 & 68.8   & 57.3 & 65.7    \\ \cdashline{2-10}
 & \textbf{MOSAD}& \textbf{65.6} &  \textbf{74.6}   & \textbf{68.3}& \textbf{78.3}     & \textbf{68.3}& \textbf{78.3}   & \textbf{68.1}& \textbf{77.6}    \\ \hline
 \multirow{2}{*}{STD} & NCAD  & 2.4& 2.2    & 2.7& 2.2      & 4.3& 3.2    & 2.1& 2.1     \\ \cdashline{2-10}
     & \textbf{MOSAD} & \textbf{0.5}& \textbf{1.2}    & \textbf{0.8}& \textbf{0.6}      & \textbf{0.9}& \textbf{0.7}    & \textbf{1.1}& \textbf{0.7}     \\ \hline
   \multicolumn{10}{c}{$\eta \in \{18,21,24,27,30 \}$} \\ \hline
\multirow{2}{*}{Mean} & NCAD  & 64.2& 74.3   & 69.5& 78.3     & 61.8& 73.6   & 60.3& 67.8    \\ \cdashline{2-10}
  & \textbf{MOSAD} & \textbf{67.4}& \textbf{76.9}   & \textbf{73.6}& \textbf{82.2}     & \textbf{69.6}& \textbf{79.6}   & \textbf{68.2}& \textbf{78.0}    \\ \hline
\multirow{2}{*}{STD} & NCAD  & 3.5& 2.4    & 5.5& 4.7      & 2.8& 1.4    & 3.3& 2.7     \\ \cdashline{2-10}
  & \textbf{MOSAD} & \textbf{1.0}& \textbf{0.7}    & \textbf{4.2}& \textbf{3.0}      & \textbf{1.1}& \textbf{0.8}    & \textbf{0.8}& \textbf{0.9} \\ \hline \hline       
\end{tabular}
}
\label{tab:result_eta}
\end{table}

\begin{figure}[h]
    \centering
    \includegraphics[width=0.60\linewidth]{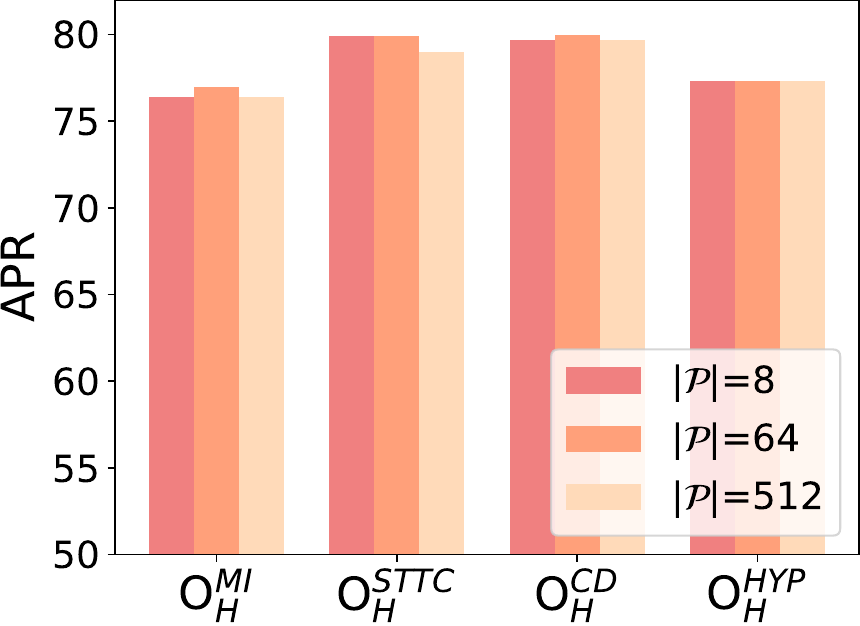}
    \includegraphics[trim={0 0.2cm 0 0 },clip,width=0.31\linewidth]{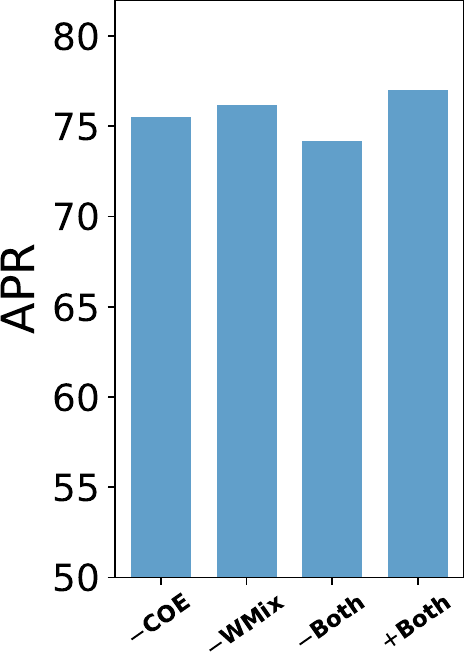}
    \caption{(\textbf{Left}) $|\mathcal{P}|$ versus \g{APR} under \textbf{O}$^{\text{Class}}_H$ on PTB-XL. (\textbf{Right}) Anomaly augmentation techniques versus APR under \textbf{O}$^{\text{MI}}_H$.} 
    \label{fig:sensitivity_p}
\end{figure}

\g{\textbf{Effect of $|\mathcal{P}|.$} Figure \ref{fig:sensitivity_p} (Left) shows the sensitivity of MOSAD to $|\mathcal{P}|,$ where $|\mathcal{P}| = \{8, 64, 512\},$ while other hyperparameters are fixed to the values presented in Section \ref{section:implement}. Based on the results from  \textbf{O}$_{H}^{\text{MI}},$ \textbf{O}$_{H}^{\text{STTC}},$ \textbf{O}$_{H}^{\text{CD}},$ and \textbf{O}$_{H}^{\text{HYP}},$ the APR scores of MOSAD varies little w.r.t. $|\mathcal{P}|.$ This indicates that MOSAD is robust against changes in} hyperparameter $|\mathcal{P}|,$ as long as $|\mathcal{P}|$ is sufficiently large.

\subsubsection{Effect of Anomaly Augmentation Techniques}

Figure \ref{fig:sensitivity_p} (Right) shows the effect of anomaly augmentation techniques on MOSAD's performance under \textbf{O}$^{\text{MI}}_H$. Note that the same anomaly augmentation techniques are also always applied to all supervised comparison methods \s{as well }under \textbf{O}$_G$ and \textbf{O}$^{\text{Class}}_H.$  -COE, -WMix, and -Both, respectively, denote that COE, WMix, or both techniques are not applied. On the other hand, +Both indicates the usage of both augmentation methods. The results indicate that the inclusion of both anomaly augmentation techniques improves the APR of MOSAD. This suggests that the synthetic anomalies assist in addressing the challenge of the limited availability of labeled anomalies.


\section{Conclusion}
While open-set anomaly detection has witnessed significant advancements in computer vision, it remains unaddressed in the time-series domain. This paper presents the first algorithm to directly tackle the open-set TSAD problem, achieving superior performance compared to both SOTA supervised and STOA unsupervised TSAD algorithms. Our proposed algorithm, MOSAD, addresses the challenges of open-set mutivariate time-series anomaly detection by its novel architecture that includes a shared Feature Extractor for temporal representation learning, a Generative head for modeling both inter-sensor and temporal patterns in normal data, a Discriminative head capable of establishing a decision boundary with only a few seen anomalies, and an Anomaly-Aware Contrastive head that uses a novel contrastive learning framework tailored to open-set TSAD to augment the feature space for improved anomaly detection performance. Our extensive experimental results not only establishes MOSAD as the new SOTA in the field, but pave the way for improved real-world applications of anomaly detection in diverse domains, such as the detection of fraud and attacks in internet networks, or the identification of pathologies in biomedical signals.

\begin{ack}
\br{The authors wish to acknowledge the financial support of the Natural Sciences and Engineering Research Council of Canada (NSERC), Fonds de recherche du Québec (FRQNT), and the Department of Electrical and Computer Engineering at McGill University. This research was enabled in part by Calcul Quebec and the Digital Research Alliance of Canada.}
\end{ack}



\bibliography{mybibfile}

\end{document}